\documentclass[letterpaper]{article} 
\usepackage{aaai2026}  
\usepackage{times}  
\usepackage{helvet}  
\usepackage{courier}  
\usepackage[hyphens]{url}  
\usepackage{graphicx} 
\usepackage{pifont}
\newcommand{\CIRCLE}{\ding{108}}
\newcommand{\LEFTcircle}{\ding{119}}
\newcommand{\Circle}{\ding{109}}
\usepackage{natbib}  
\usepackage{caption} 
\usepackage{algorithm}
\usepackage{algorithmic}
\usepackage{acronym}
\usepackage{longtable}
\usepackage{booktabs}
\usepackage[inline]{enumitem}
\usepackage{csquotes}
\usepackage{newfloat}
\usepackage{listings}

\usepackage{xcolor}
\newif\ifmarkup
\markupfalse   
\ifmarkup
  \newcommand{\alessiocut}[1]{{\color{red}\sout{#1}}}
  \newcommand{\alessiocutlong}[1]{{\color{red}[CUT: #1]}}
  
  \newcommand{\alessionote}[1]{{\color{orange}\footnotesize [NOTE: #1]}}
\else
  \newcommand{\alessiocut}[1]{}
  \newcommand{\alessiocutlong}[1]{}
  
  \newcommand{\alessionote}[1]{}
\fi

\DeclareCaptionStyle{ruled}{labelfont=normalfont,labelsep=colon,strut=off} 
\floatstyle{ruled}
\newfloat{listing}{tb}{lst}{}
\floatname{listing}{Listing}
\begin{document}

\title{The AI Assessment Sandbox Configurator: A Framework to Support Technical Assessment in AI Regulatory Sandboxes}

\author{
    Alessio Buscemi\textsuperscript{1},
    German Castignani\textsuperscript{1},
    Daniele Pagani\textsuperscript{1},
    Maxime Cordy\textsuperscript{2},
    Jordi Cabot\textsuperscript{1,2}
}
\affiliations{
    \textsuperscript{1}Luxembourg Institute of Science and Technology (LIST), Esch-sur-Alzette, Luxembourg\\
    \textsuperscript{2}University of Luxembourg, Esch-sur-Alzette, Luxembourg\\
    \{name.surname\}@list.lu, \{name.surname\}@uni.lu
}

\maketitle

\begin{abstract}
The EU's Artificial Intelligence Act requires all Member States to establish AI Regulatory Sandboxes (AIRS) by August 2027: supervised environments bringing together national Competent Authorities, technical experts, and the organisations under assessment. When AIRS engagements include structured technical testing, running such testing at scale demands dedicated infrastructure, yet the tooling ecosystem remains structurally fragmented, with heterogeneous tools producing outputs that are difficult to compare, trace, and reuse. From the procedural conditions of AIRS engagements and the AI Act obligations for high-risk systems, we derive 11 architectural and governance requirements for the infrastructure that operationalises technical testing within an AIRS. In response to these requirements, we introduce the AI Assessment Sandbox Configurator, an open-source framework combining a curated Catalogue of tests and controls accessed through a stable plug-in API, a shared data model that harmonises heterogeneous outputs, role-specific dashboards for multi-disciplinary interpretation, and audience-segmented reporting. We describe the architecture and current release, and report an early-stage pilot that exercised the harmonisation and reporting layers within a live AIRS engagement and contributed to an official Exit Report. We discuss the roadmap, the governance questions raised by the Catalogue's tiered contribution model, and the institutional pathways through which an open-source assessment ecosystem could emerge across Member States.
\end{abstract}

\begin{links}
    \link{Code}{https://github.com/lux-ai-factory/aisc}
\end{links}

\acrodef{AI}{Artificial Intelligence}
\acrodef{FCA}{Financial Conduct Authority}
\acrodef{AI Act}{Artificial Intelligence Act}
\acrodef{GDPR}{General Data Protection Regulation}
\acrodef{CRA}{Cyber Resilience Act}
\acrodef{SME}{Small-medium enterprise}
\acrodef{GPAI}{General Purpose AI}
\acrodef{DGA}{Data Governance Act}
\acrodef{DSA}{Digital Services Act}
\acrodef{DMA}{Digital Markets Act}
\acrodef{AIRS}{AI Regulatory Sandbox}
\acrodefplural{AIRS}[AIRS]{AI Regulatory Sandboxes}
\acrodef{EUSAIR}{EU Regulatory Sandboxes for AI}
\acrodef{EDIH}{European Digital Innovation Hub}
\acrodef{TEF}{Testing and Experimentation Facilities}
\acrodef{EuroHPC}{European High-Performance Computing}
\acrodef{EDIC}{European Digital Infrastructure Consortia}
\acrodef{AIoD}{AI-on-Demand Platform}
\acrodef{AI HLEG}{High-Level Expert Group on Artificial Intelligence}
\acrodef{AI RMF}{AI Risk Management Framework}
\acrodef{CA}{Competent Authority}
\acrodefplural{CA}[CAs]{Competent Authorities}
\acrodef{NIST}{National Institute of Standards and Technology}
\acrodef{RMF}{Risk Management Framework}
\acrodef{CEN}{European Committee for Standardization}
\acrodef{CENELEC}{European Committee for Electrotechnical Standardization}
\acrodef{JTC}{Joint Technical Committee}
\acrodef{RAG}{Retrieval Augmented Generation}
\acrodef{NLP}{Natural Language Processing}
\acrodef{HPC}{High Performance Computing}
\acrodef{DSL}{Domain-Specific Language}
\acrodef{RISE}{Research Institutes of Sweden}
\acrodef{NFR}{requirement}
\acrodef{EC}{European Commission}
\acrodef{OECD}{Organisation for Economic Co-operation and Development}
\acrodef{API}{Application Programming Interface}
\acrodef{BESSER}{Building bEtter Smart Software fastER}
\acrodef{MAS}{Multi-Agent System}
\acrodef{LLM}{Large Language Model}
\acrodef{MLOps}{Machine Learning Operations}
\acrodef{RBAC}{Role-Based Access Control}
\acrodef{ORM}{Object-Relational Mapping}
\acrodefplural{LLM}[LLMs]{Large Language Models}
\acrodef{DBMS}{Database Management System}
\acrodef{RFC}{Request for Comments}
\acrodef{UI}{User Interface}
\acrodef{AIP}{AI Provider}
\acrodef{TE}{Technical Expert}
\acrodef{DPIA}{Data Protection Impact Assessment}
\acrodef{LIST}{Luxembourg Institute of Science and Technology}

\section{Introduction}
\label{sec:introduction}
AI systems are increasingly deployed in critical sectors where failures may have serious societal consequences \cite{bengio2024managing, wang2022artificial}, prompting regulatory efforts to ensure trustworthiness, fairness, and transparency \cite{smuha2021race, novelli2024taking, huang2024ai}. The EU's \ac{AI Act} \cite{eu_ai_act_2024} is among the most comprehensive responses, introducing a risk-based framework with ex ante assessment and ongoing oversight for high-risk systems, with obligations ranging from provider self-assessment, through third-party conformity assessment, to supervised testing within \acp{AIRS} --- environments where providers test systems for performance, risks, and compliance in collaboration with \acp{CA} and domain experts prior to market deployment. All EU Member States must establish at least one \ac{AIRS} by August 2027, and analogous environments are emerging globally \cite{moraes2025regulatory, nabil2024artificial, cahane2025nascent, aiactionplan}.
When AIRS engagements include structured technical testing, they impose conditions that no single existing tool was designed to meet. Multiple parties with different roles and access rights must collaborate within a supervised regulatory environment, and the \ac{CA} requires evidence it can rely upon for the Exit Report and any subsequent regulatory follow-up. Article~58(2)(i) adds a further condition: in tasking the AI Office with collecting aggregated sandbox findings, it presupposes a degree of structure and comparability across engagements that the current fragmented tooling ecosystem does not deliver. Together, these conditions demand dedicated infrastructure, which we refer to as an \emph{AI Assessment Sandbox}.
The assessment tooling ecosystem compounds the problem: outputs resist harmonisation across heterogeneous open-source, proprietary, and bespoke tools \cite{portegies2025decoding, liang2022helm}, metrics and thresholds vary across sectors and risk levels, and existing tools offer limited support for structured multi-disciplinary collaboration. This fragmentation is a structural obstacle to the cross-engagement learning Article~58(2)(i) anticipates, and it shapes the requirements any AI Assessment Sandbox must satisfy.

We make four contributions. First, we identify 11 architectural and governance requirements for AI Assessment Sandboxes, derived from the procedural conditions that Articles 57 and 58 place on AIRS engagements involving technical testing (\emph{digital sandboxing}, Section~\ref{sec:background}), from consultations with Competent Authorities and complementary European initiatives, and from the authors' experience as technical testers in the AIRS pilot reported in Section~\ref{sec:pilot}. Second, we describe the \emph{AI Assessment Sandbox Configurator} (hereafter, the Configurator), an open-source framework implementing these requirements: configuration components capture the AI system's profile and assessment intent, and a runtime sandbox ``\`{a} la carte'' is instantiated from that configuration, with all components substitutable except two stable interfaces (the Catalogue plug-in API and the shared data model). Third, we present an early-stage pilot exercising the harmonisation and reporting layers under real regulatory conditions and contributing to an official Exit Report. Fourth, we discuss the Configurator's scope, the governance questions raised by the Catalogue's tiered contribution model, and the institutional pathways through which an open-source assessment ecosystem could emerge across Member States.

\section{Background}
\label{sec:background}

The \ac{AI Act} \cite{almadaEUAIAct2025} classifies AI systems into minimal, limited, high, and unacceptable risk categories, with the strictest obligations on high-risk systems, and reflects the seven Trustworthy AI requirements articulated by the \ac{AI HLEG} \cite{eu_trustworthy_ai_2019}.

Regulatory uncertainty has emerged as a practical concern for companies operating under the \ac{AI Act}, with empirical studies and industry surveys indicating that interpretive ambiguity contributes to delayed investment, uneven compliance readiness, and fragmented risk management \cite{lewis2025regulatorylearning, deloitte2024ai, dlapiper2025pause, draghi2024competitiveness, arnal2024implementation, regulatinguncertainty2025}. This underscores the need for operational mechanisms that translate legal requirements into verifiable technical practices.
To this end, the \ac{AI Act} complements its risk-based obligations with instruments aimed at regulatory learning and innovation support \cite{schrepelAdaptiveRegulation2025, lewis2025regulatorylearning, deckenbrunnen2026bathtub}, among which \acp{AIRS} play a central role \cite{longo2025unboxing, povscic2022regulatory}. Article~57 mandates that each Member State establish at least one \ac{AIRS}, with voluntary participation and free access to core activities for startups and SMEs. Within an AIRS, a \ac{CA} provides regulatory guidance, clarifies obligations, and gives feedback on system design, risk management, and conformity pathways. Unlike conformity assessment, \acp{AIRS} create a pre-commercial environment where the resulting \emph{Exit Report} can later support audits and foster compliance-by-design, legal certainty for grey-zone technologies, and cross-border collaboration towards a coherent EU AI market \cite{buocz2023regulatory, yordanova2024regulating}. The importance of technical testing within \acp{AIRS} is recognised in Article~58(2)(i) and in the current draft of the Implementing Act on AIRS \cite{implementingAct}.

In practice, AIRS engagements vary considerably. \ac{EUSAIR}, the EU-funded project coordinating pilots across Member States ahead of the entry into force of the obligation, observes a useful distinction between engagements centred on legal and procedural oversight and those that additionally involve structured technical testing producing evidence as requested by the \ac{CA}. The latter is referred to as a \emph{digital sandbox} \cite{eusair2025}. This is a descriptive distinction, not a formal legal category, and does not prescribe how the technical testing component should be implemented. We use \emph{AI Assessment Sandbox} to refer to the infrastructure that operationalises the digital-sandbox component of such an engagement.
The tooling ecosystem remains fragmented. A recent analysis of ethical AI tools finds significant imbalance in the principles they address and a strong reliance on self-assessment \cite{portegies2025decoding}; similar concerns arise in AI auditing, where the field still lacks shared terminology and stable standards for what counts as an audit and how results should support accountability \cite{mokander2023auditing}.

\textbf{Experiment tracking platforms} such as MLflow \cite{zaharia2018accelerating} support metadata logging and lineage tracking within individual development pipelines, focusing on experiment management within a single ML project.

\textbf{Model documentation frameworks} such as model cards \cite{mitchell2019} and datasheets for datasets \cite{gebru2021} introduced structured templates for disclosing model performance characteristics and dataset provenance: manually authored, human-readable artefacts intended for communication rather than downstream integration.

\textbf{Comprehensive evaluation frameworks} such as HELM \cite{liang2022helm} evaluate language models across multiple dimensions (accuracy, robustness, fairness, bias, toxicity) under standardised conditions within a single integrated framework.

\textbf{National AI assessment tools} such as AI Verify \cite{aiverify2023} provide governance frameworks and testing toolkits combining automated tests with governance process checks.

\textbf{Reference architectures} such as PARMA \cite{pintz2024parma} specify the structure of multi-party, reproducible AI trustworthiness assessments without prescribing the assessment tools, data formats, or deployment infrastructure. This generality maximises flexibility but leaves the concrete instantiation to the adopting organisation.

\textbf{Regulation-derived benchmark suites} such as COMPL-AI \cite{complai2024} map AI Act
obligations onto technical requirements and executable tests: 18 requirements and 27 LLM
benchmarks in an open-source suite. COMPL-AI is the closest work in regulatory
grounding, but it prescribes a fixed benchmark set for language models rather
than supporting the heterogeneous tools and multi-party process an AIRS
engagement requires.

Our survey covers tools whose public documentation is sufficient to assess them against R1--R11; it is not an exhaustive map of the compliance-tooling market, and in particular excludes proprietary offerings whose internal mechanisms are not publicly documented.

\section{Requirements}
\label{sec:requirements}

The requirements below were derived in three stages. We first elicited an initial set from the procedural conditions imposed by AIRS digital sandboxing as set out in Articles 57 and 58 of the AI Act. We then refined the set against the operational expectations expressed in consultations with Competent Authorities and with the European initiatives whose mandates intersect with AIRS --- namely Testing and Experimentation Facilities (TEFs), AI Factories, and European Digital Innovation Hubs (EDIHs), whose importance for AIRS is acknowledged in the current draft of the Implementing Act --- conducted in the context of the funding initiative supporting this work. Finally, we refined the set further against the authors' direct experience as technical testers in an AIRS pilot coordinated by EUSAIR, in particular the interactions with the CA during the engagement reported in Section~\ref{sec:pilot}. The \emph{Origin} column of Table~\ref{tab:requirements} records which of these three stages each requirement derives from.

AIRS digital sandboxing imposes a specific and demanding set of conditions: multiple parties collaborate within a supervised regulatory environment, each with different roles, access rights, and interpretive needs; the \ac{CA} requires tamper-evident evidence for the Exit Report and any subsequent audit; findings must be machine-readable and comparable across engagements; privacy-sensitive sectors require assessment to run wholly within the provider's environment; and no single tool covers the full range of trustworthiness dimensions the \ac{AI Act} encompasses.
These conditions translate into requirements for the infrastructure that supports technical testing. Table~\ref{tab:requirements} provides the mapping to \ac{AI Act} articles. All 11 have direct relevance to one or more provider obligations set out in Articles~9 to 15 for high-risk systems.
\begin{table*}[tp]
\footnotesize
\begin{tabular}{p{2.3cm} p{10.2cm} p{2.35cm} p{1.05cm}}
\hline
\textbf{Requirement} & \textbf{Description} & \textbf{Relevant AI Act Articles} & \textbf{Origin}\\
\hline
\textbf{R1: Customisability and Modularity} & Each engagement involves a distinct system, sector, and risk profile, with metrics, thresholds, and testing methods that cannot be prescribed in advance. The sandbox must support ``\`{a} la carte'' configuration of tests, metrics, and pipelines, tailoring every engagement to its context rather than forcing it into a generic protocol. & Art.~9 (Risk Mgmt), Art.~15 (Accuracy) & A, P\\
\hline
\textbf{R2: Compatibility with External Assessment Catalogue} & No single tool covers the full range of trustworthiness dimensions a \ac{CA} may require evidence on. The sandbox must integrate heterogeneous external tools, benchmarks, and datasets, enabling unified execution and result harmonisation regardless of original format, so the evidence base reflects the full scope of the engagement. & Art.~10 (Data Governance), Art.~13 (Transparency), Art.~15 (Accuracy) & A, C\\
\hline
\textbf{R3: Visual Pipelines and Tailored Dashboards} & A \ac{CA} official, a technical tester, a legal expert, and a product manager cannot all be served by the same view of assessment results. The sandbox must provide role-specific, real-time dashboards through which each group interprets harmonised evidence in the terms of its regulatory, technical, or governance responsibility, alongside low-code interfaces for composing and monitoring pipelines. & Art.~11 (Documentation), Art.~13 (Transparency), Art.~14 (Human Oversight) & C, P\\
\hline
\textbf{R4: Open-Source Core} & For \ac{CA} oversight to be meaningful, the assessment infrastructure itself must be inspectable; a proprietary black-box sandbox undermines the transparency that supervised regulatory testing requires. The core codebase must be released under a permissive licence, enabling community scrutiny, co-development by regulators and industry, and alignment with European digital sovereignty objectives. & Art.~11 (Documentation), Art.~13 (Transparency) & C\\
\hline
\textbf{R5: Plug-in Architecture} & The \ac{AI Act} spans sectors and system types with domain-specific testing needs that no single team can anticipate. The sandbox must expose stable APIs enabling third parties to add sector-specific tests, visualisers, and data connectors without modifying the core, so the assessment ecosystem grows alongside regulatory needs. & Art.~15 (sector-specific accuracy and robustness) & A, C\\
\hline
\textbf{R6: Deployment Portability} & Providers in privacy-sensitive sectors such as healthcare or finance cannot allow assessment data to leave their control. The sandbox must be deployable on-premises or in sovereign clouds, producing identical artefacts in any deployment context, so portability constraints do not compromise cross-engagement comparability. & Art.~15 (Robustness) & C\\
\hline
\textbf{R7: Role-based Access Control} & Multi-party engagements create strict information boundaries: the provider cannot access the \ac{CA}'s internal deliberations; the \ac{CA} may not be entitled to proprietary model details; testers may have no business reviewing legal risk assessments. The sandbox must enforce fine-grained privileges, separation of duties, and data-zone isolation. & Art.~10 (Data Governance), Art.~15 (Cybersecurity) & A, P\\
\hline
\textbf{R8: Automated Report Generation} & The Exit Report must communicate findings to the provider, the \ac{CA}, and potentially affected communities from one evidence base. The sandbox must generate machine- and human-readable reports automatically, with audience-segmented sections, in a format supporting aggregation into a cross-engagement evidence base at national and European scale. & Art.~11 (Documentation), Art.~12 (Record-keeping) & A, P \\
\hline
\textbf{R9: Immutable Audit Trail} & Evidentiary value depends on the \ac{CA}'s ability to verify that results were not altered between execution and reporting. The sandbox must maintain a tamper-evident log of data ingress, code versions, parameters, and results, exportable to the \ac{CA}'s traceability layer and able to withstand post-hoc audit. & Art.~12 (Record-keeping), Art.~15 (Cybersecurity) & A, C \\
\hline
\textbf{R10: Shared Data Model and Persistent Storage} & The cross-engagement aggregation Article~58(2)(i) anticipates is feasible only if results from heterogeneous tools are stored in a common, machine-readable format. The sandbox must integrate a \ac{DBMS}-agnostic backend built on a shared data model harmonising all outputs, automated and manual, into one repository ensuring traceability, reproducibility, and comparability. & Art.~10 (Data Governance), Art.~12 (Record-keeping) & A\\
\hline
\textbf{R11: Cybersecurity and Threat Mitigation} & The sandbox handles model endpoints, evaluation datasets, and compliance evidence for systems under regulatory scrutiny, making it a high-value target. Industry-grade security controls must be applied throughout the assessment lifecycle to prevent unauthorised access, detect anomalies, and protect the integrity of the evidence on which \ac{CA} decisions rest. & Art.~10 (Data Governance), Art.~15 (Cybersecurity) & A, C\\
\hline
\end{tabular}
\caption{Requirements for an AI Assessment Sandbox grounded in the conditions of AIRS digital sandboxing, and their mapping to AI Act articles. \emph{Origin} records the derivation stage (Section~\ref{sec:requirements}): A = Articles 57--58; C = consultations with CAs and complementary European initiatives; P = the authors' experience as technical testers in the pilot of Section~\ref{sec:pilot}.}
\label{tab:requirements}
\end{table*}
Three of these requirements have a particularly direct architectural consequence. R1 rules out infrastructure that prescribes a fixed evaluation protocol; R3 rules out infrastructure that serves all stakeholders through a single undifferentiated interface; R6 rules out infrastructure that cannot be deployed entirely within the provider's environment. Taken together, they push against a single, fixed AI Assessment Sandbox, which could satisfy each of them only at the cost of restricting the sectors it serves (R1), baking in role structures that differ across organisations (R3), or falling back on a lowest-common-denominator deployment (R6). A more flexible response is one in which each sandbox is configured from a description of the specific engagement, its stakeholder structure, and its reporting needs. We do not claim this is the only viable architectural response, but it is the one we adopt, and Section~\ref{sec:architecture} describes how the Configurator implements it.

Table~\ref{tab:related_work} assesses the tools surveyed in Section~\ref{sec:background} against the 11 requirements, alongside a self-assessment of the Configurator. We adopt the following criteria: \CIRCLE{} indicates the tool, as released and documented, addresses the requirement substantively; \LEFTcircle{} indicates partial coverage, either because the mechanism exists but is not yet fully exercised in deployment, or because the tool addresses the requirement only within a restricted scope; \Circle{} indicates that the requirement is outside the tool's design scope or is not addressed.

MLflow provides genuine modularity, lineage tracking, and deployment portability within single-pipeline experiment management, but is not designed for heterogeneous external-tool integration or multi-party regulatory contexts. Model cards and datasheets are conceptual precursors to structured multi-audience disclosure but as manually produced documentation artefacts they do not address infrastructure-level requirements. HELM delivers depth on evaluation scenarios, metrics, and result schemas within its own framework, with genuine extensibility points, but as a single integrated benchmark it cannot accommodate the heterogeneous, sector-tailored testing AIRS engagements require. AI Verify is the closest in purpose, offering structured assessment with governance process integration, configurable test selection, and automated report generation; we rate it \Circle{} on R2 because its testing toolkit is tightly coupled to its framework, with no documented mechanism for harmonising outputs from arbitrary external tools, and it is not designed for the kind of cross-engagement aggregation Article~58(2)(i) anticipates. PARMA correctly identifies the architectural shape of the solution as multi-party, reproducible, and tool-agnostic; operating at this abstract layer maximises conceptual flexibility but leaves concrete integration, harmonisation, and deployment to the adopting organisation, and we mark R6 as not applicable since PARMA is a reference architecture rather than a deployable system. COMPL-AI is the most regulatorily grounded of the tools surveyed, deriving its benchmark set directly from a technical interpretation of the Act, and its ratings largely track HELM's for the same reason: depth within its own suite, with community contribution routed through the repository rather than a stable third-party API (R5). We rate it \Circle{} on R2 on the same grounds as HELM and AI Verify, and \Circle{} on R11 because it benchmarks the cybersecurity of the models under test rather than providing security controls for the assessment infrastructure itself.

For the Configurator's own row, R2, R3, and R8 are rated \CIRCLE{} as architecturally committed and partially exercised in the pilot reported in Section~\ref{sec:pilot}. R1, R4, R5, R6, and R7 are rated \CIRCLE{} as architecturally committed but not exercised in the pilot. R9, R10, and R11 are rated \LEFTcircle{} for a shared reason: their assurance depends substantially on the environment in which the framework is deployed, not on the framework alone. The shared data model is in active use and validated against the current plugin set; tamper-evident logging via ImmuDB is operational for two components; and the framework provides RBAC, credential isolation, and authenticated inter-component channels. In each case, however, durable assurance also rests on the hosting environment: storage backend and retention policy for R10, clock synchronisation and custody of the published root hash for R9, and network segmentation, identity provider integration, secrets management, patching, and monitoring for R11. Section~\ref{sub:limitations} details the current coverage.

\begin{table*}[!t]
\footnotesize
\begin{tabular}{p{5cm} p{0.6cm} p{0.6cm} p{0.6cm} p{0.6cm} p{0.6cm} p{0.6cm} p{0.6cm} p{0.6cm} p{0.6cm} p{0.6cm} p{0.6cm}}
\hline
\textbf{Tool} & \textbf{R1} & \textbf{R2} & \textbf{R3} & \textbf{R4} & \textbf{R5} & \textbf{R6} & \textbf{R7} & \textbf{R8} & \textbf{R9} & \textbf{R10} & \textbf{R11} \\
\hline
MLflow & \LEFTcircle & \Circle & \LEFTcircle & \CIRCLE & \LEFTcircle & \CIRCLE & \Circle & \Circle & \LEFTcircle & \LEFTcircle & \LEFTcircle \\
Model cards / Datasheets & \LEFTcircle & \Circle & \Circle & \CIRCLE & \Circle & \Circle & \Circle & \LEFTcircle & \Circle & \Circle & \Circle \\
HELM & \LEFTcircle & \Circle & \LEFTcircle & \CIRCLE & \LEFTcircle & \CIRCLE & \Circle & \LEFTcircle & \LEFTcircle & \LEFTcircle & \Circle \\
AI Verify & \LEFTcircle & \Circle & \LEFTcircle & \CIRCLE & \LEFTcircle & \CIRCLE & \LEFTcircle & \LEFTcircle & \LEFTcircle & \LEFTcircle & \LEFTcircle \\
PARMA & \CIRCLE & \LEFTcircle & \LEFTcircle & \CIRCLE & \CIRCLE &\CIRCLE & \LEFTcircle & \LEFTcircle & \LEFTcircle & \LEFTcircle & \Circle \\
COMPL-AI & \LEFTcircle & \Circle & \LEFTcircle & \CIRCLE & \LEFTcircle & \CIRCLE & \Circle & \LEFTcircle & \Circle & \LEFTcircle & \Circle \\
\hline
\textbf{Configurator} & \CIRCLE & \CIRCLE & \CIRCLE & \CIRCLE & \CIRCLE & \CIRCLE & \CIRCLE & \CIRCLE & \LEFTcircle & \LEFTcircle & \LEFTcircle \\
\hline
\end{tabular}
\caption{Assessment of related tools against the 11 requirements. \CIRCLE~= addressed substantively; \LEFTcircle~= partially addressed or addressed within restricted scope; \Circle~= not addressed or out of design scope. Configurator self-ratings are deliberately conservative where features are architecturally committed but not yet exercised at scale.}
\label{tab:related_work}
\end{table*}

\section{The AI Assessment Sandbox Configurator}
\label{sec:architecture}

This section describes the current release. Roadmap components are consolidated in Section~\ref{sub:limitations} and flagged as such wherever they are referenced below.

\begin{figure*}[!t]
\centering
\includegraphics[width=\textwidth]{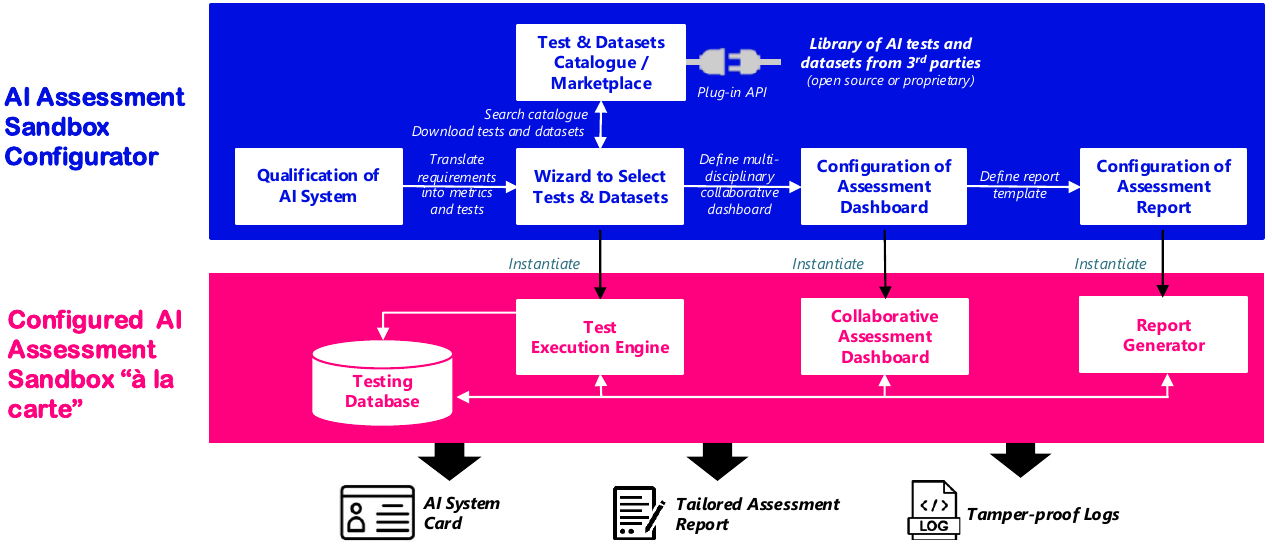}
\caption{Architecture of the Configurator. The upper blue zone (\emph{AI Assessment Sandbox Configurator}) contains the configuration-time components; the lower pink zone (\emph{Configured AI Assessment Sandbox ``\`{a} la carte''}) contains the runtime components instantiated once configuration is complete. The Catalogue (discovery layer) and Testing Database (harmonisation layer) provide the two stable interfaces; all other components are substitutable.}
\label{fig:architecture}
\end{figure*}

The Configurator is an open-source framework (Apache 2.0 licence)\footnote{Source code, plug-in API documentation, and the \ac{RFC} process: \url{https://github.com/lux-ai-factory/aisc}.} developed within a large-scale EU- and nationally-funded AI infrastructure initiative, with an ongoing public RFC process gathering input on the framework's evolution. It is packaged using Docker Compose for modular containerised deployment on any infrastructure (R6).
The architecture separates configuration-time from runtime concerns across two zones (Figure~\ref{fig:architecture}). The upper blue zone (the \emph{AI Assessment Sandbox Configurator}) contains the configuration-time components: the user describes the AI system, selects the tests and controls to apply, and configures how results will be visualised and reported. Once configuration is complete, the Configurator \emph{instantiates} the lower pink zone (the \emph{Configured AI Assessment Sandbox ``\`{a} la carte''}), a runtime environment in which tests are executed, stakeholders collaborate on results, and tailored reports are produced.
Two foundational components host the stable interfaces on which the rest of the architecture depends: the Catalogue hosts the plug-in API (discovery layer), and the Testing Database hosts the shared data model (harmonisation layer). Every other component is substitutable provided it respects these two contracts: plugins must be registered through the Catalogue, and results must be written to the shared data model. We describe these as ``stable interfaces'' rather than invariant: their internal schemas evolve under versioned governance. What is invariant is the architectural commitment that all components communicate through them.
The remainder of this section follows the data flow, from qualification of the AI system, through plugin selection and configuration, into the instantiated runtime, and finally to the outputs, with each subsection signposting the requirements (Section~\ref{sec:requirements}) it addresses.

\subsection{Qualification of the AI System}
\label{sub:qualification}

\textit{Qualification is the prerequisite for all subsequent steps; it produces the AI System Card that drives downstream plugin selection (\textbf{R2}, \textbf{R5}).}

The starting point is the \emph{AI System Qualification Form}, a structured questionnaire capturing the AI system's risk category, the types of end users and third parties affected, the deployment sector, and the trustworthiness dimensions to be assessed. The questionnaire balances granularity with brevity, and multiple versions are envisioned, tiered by company size and sector, mirroring the way the \ac{AI Act} scales its requirements to risk.
The Configurator provides a dedicated interface through which the form is completed and the resulting \emph{AI System Card}\footnote{We use ``AI System Card'' for the standardised, machine-readable system profile defined here, following the AI Cards line of work \cite{golpayeganiAICardsApplied2024a}. The same term has been used elsewhere in industry for a different artefact documenting deployed systems for end users; the two are unrelated.}, a standardised profile of the system, is generated. In the current release, the card is inferred from the form using a single \ac{LLM} call with a structured prompt.

We are explicit about the limitations. A non-deterministic LLM call sits at the start of a workflow whose outputs may eventually feed regulatory artefacts. Failure modes identified during internal development use, rather than through systematic evaluation, include occasional misclassification of sector, under-specification of affected stakeholder groups, and inconsistent verbosity across runs on similar inputs. We have not characterised run-to-run variance quantitatively, and we do not claim a measured error rate for the generator; a systematic consistency evaluation is a near-term roadmap item (Section~\ref{sub:limitations}).
The user retains the ability to manually amend the card before it is used downstream, and during internal exercises this step has been used to correct each of these failure modes. A future release will replace the single LLM call with an agentic system performing inference with formal verification against the AI System Card schema; manual amendment remains available in all releases, ensuring human oversight. The card can be downloaded as PDF for human inspection or JSON for machine readability.

The AI System Card serves two purposes. First, it \emph{drives plugin selection} from the Catalogue (\S\ref{sub:catalogue}). Second, as a machine-readable artefact in a standardised format, it \emph{enables cross-engagement comparison}: if widely adopted, AI System Cards would facilitate sharing of qualification experiences across jurisdictions and deployments, supporting AI Office oversight at national and European levels \cite{golpayeganiAICardsApplied2024a}.

\subsection{Catalogue of AI Tests and Controls}
\label{sub:catalogue}

\textit{Addresses \textbf{R2} (compatibility with external assessment catalogues) and \textbf{R5} (plug-in architecture).}

The Catalogue is a structured, extensible library of testing tools, controls checklists, and datasets supporting both automated and manual assessment, against which the AI System Card is matched to identify relevant plugins. At the time of writing, it is available as a public website where anyone can browse available solutions and submit new ones. A downloadable, open-source version is planned, enabling organisations to install a local instance and register private or proprietary plugins without exposing them publicly. The long-term vision is a community-curated index of assessment services specialised by vertical domain. The comparison with general-purpose model hubs only goes so far: assessment tooling follows regulatory rather than developer incentives, so the contributor base will be narrower and more institutional --- a difference the tiered governance below reflects.

A plugin is fully integrated when four conditions, which together form the \emph{integration contract}, are met: the \emph{hooks} defined by the plugin interface are implemented; the plugin's \emph{output mapping} to the shared data model (Section~\ref{sub:domain_model}) is defined; plugin-specific \emph{reporting} logic and templates are implemented; and \emph{documentation} follows a standardised metadata template. This contract is enforced through a plug-in API, the stable interface through which plugins are registered, validated, and made available; the interface signatures, including optional hooks for visualisation metadata, Catalogue presentation, and dataset-driven configuration inference, are documented in the project repository (Section~\ref{sec:architecture}). This ensures every plugin can be executed uniformly, its results stored comparably, and its outputs surfaced coherently in dashboards and reports, regardless of the original tool's output format or methodology.
At the time of writing, 10 plugins are integrated against the four conditions above, across two trustworthiness domains: \ac{LLM} safety and robustness (bias detection, multilingual performance, jailbreak resistance, \ac{RAG} evaluation, agentic safety), and classical ML performance (classification, regression, anomaly detection, data drift, model explainability). For each plugin we verified that (i) outputs could be expressed in the shared data model without lossy coercion, (ii) the resulting records were retrievable and aggregable through the Testing Database \ac{API}, and (iii) downstream dashboards and report sections rendered the harmonised data without plugin-specific logic. This validation establishes the data model's adequacy for the current set; it does not establish adequacy for novel domains, and Section~\ref{sub:limitations} treats this as an open item.
To scale through community contributions while preserving the trustworthiness of its core, the Catalogue follows a three-tier plugin model (Table~\ref{tab:plugin_tiers}), mirroring mature open-source ecosystems. Only the Core tier is operational at the time of writing; the Verified and Community tiers are part of the longer-term roadmap and are discussed in Section~\ref{sec:discussion}.

\begin{table*}[!t]
\footnotesize
\begin{tabular}{p{2.3cm} p{2.9cm} p{5.35cm} p{5.35cm}}
\hline
\textbf{Tier} & \textbf{Location} & \textbf{Governance} & \textbf{Trust \& Visibility} \\
\hline
\textbf{Core} \textit{(active)} & Main project repository & Project team; formally supported and maintained & Highest trust; reference implementations \\
\hline
\textbf{Verified} \textit{(planned)} & Author's own repository & Verification badge issued by project after passing published checklist & Endorsed\\
\hline
\textbf{Community} \textit{(planned)} & Author's own repository & Self-registered; unverified label & Listed for discoverability; no implied endorsement \\
\hline
\end{tabular}
\caption{Three-tier governance model for Catalogue plugins. Only the Core tier is currently operational; the Verified and Community tiers are part of the longer-term roadmap (Section~\ref{sec:discussion}).}
\label{tab:plugin_tiers}
\end{table*}

Contributing a plugin requires structured metadata (name, description, licence, tier declaration, provider, project hyperlink, version, project maturity level, optional scientific reference, verification targets, sector, target systems, and optionally a declaration of which legal requirements the solution addresses); a representative input specification with examples; documentation of all configuration parameters; a full output specification covering each metric's data type, range, unit, derivation formula where applicable, and contextual guidance; preferred visualisation types; and guidance on report presentation. This ensures plugins are not merely technically integrated but interpretable and actionable by the multi-disciplinary experts who use them.
The Catalogue's test taxonomy translates the EU's seven trustworthiness dimensions \cite{eu_trustworthy_ai_2019}\footnote{Human agency and oversight; technical robustness and safety; privacy and data governance; transparency; diversity, non-discrimination and fairness; societal and environmental wellbeing; and accountability.} into concrete, measurable sub-dimensions and evaluation targets. For generative systems it is complemented by the OWASP Top~10 for \ac{LLM} Applications \cite{owasp_llm_2025}, which covers robustness and safety patterns that the seven dimensions address only at a coarse granularity. Controls are organised across a set of governance dimensions covering accountability, quality management, and risk management. Taxonomies for target systems (drawing on Hugging Face's model tagging scheme) and sectors (sourced from the \ac{OECD} catalogue of AI tools \cite{oecd2025toolkit}) round out the structure.

\subsection{Manual Selection of Tests and Controls}
\label{sub:wizard}

\textit{Like Qualification, the selection step is a precondition for downstream configuration; it operates on the Catalogue (\textbf{R2}, \textbf{R5}).}

Once the AI System Card is generated, the user selects the specific tests and controls to apply. At the time of writing, no automated assistant is available: the user navigates the Catalogue manually, using the filters exposed by its taxonomy (trustworthiness dimension, target system type, sector, verification type). This manual workflow surfaces the structure of the Catalogue itself, which is valuable while the taxonomy is being refined through the \ac{RFC} process.
Two questions arise about who performs this step. The first is identity. In the engagements we target, selection is performed by the technical tester in consultation with the provider, since it requires familiarity with both the system under assessment and the Catalogue taxonomy, while the \ac{CA} retains the authority to mandate a minimum set of tests and controls the engagement must cover. The same division applies to the configuration step in Section~\ref{sub:config_dashboard}. The second question is incentive. Wherever the provider drives or influences selection, there is an evident conflict of interest: the party whose system is under scrutiny partly determines which evidence is produced, and an assessment can be made to look favourable through omission rather than misreporting. The Configurator mitigates this conflict without resolving it. Selections are recorded in the audit trail and reproduced in the Tailored Assessment Report, so the scope of an assessment travels with its results and a \ac{CA} can see what was not tested as readily as what was. Whether selection authority should ultimately rest with the provider, the tester, or the \ac{CA} is a governance question that the framework makes visible but does not settle, and it connects directly to the vetting question in Section~\ref{sub:discussion_governance}.

\subsection{Configuration of Assessment Dashboard and Report}
\label{sub:config_dashboard}

\textit{Addresses \textbf{R1} (customisability), \textbf{R3} (role-specific dashboards), \textbf{R7} (role-based access control), and \textbf{R8} (audience-segmented report generation).}

With tests selected, the user configures how results will be presented, both during runtime and in the final report. For runtime, the user specifies per plugin which visualisations are displayed and which roles --- business process owners, regulators, technical AI experts, legal and compliance specialists, ethical experts --- may access them. For the report, the user assigns plots and results to audience-segmented sections. Multiple report templates can be defined (e.g., a confidential internal version, a public summary, and a regulator-specific dossier), enabling targeted communication from a single underlying evidence base. Selection (Section~\ref{sub:wizard}) and configuration are independent steps and need not be performed in that order.

\subsection{Instantiation of the Sandbox ``\`{a} la carte''}
\label{sub:instantiation}

\textit{Addresses \textbf{R1} (``\`{a} la carte'' configuration of every engagement to a specific system, sector, and risk profile).}

At this point, configuration is complete. The Configurator \emph{instantiates} the bespoke Assessment Sandbox (lower pink zone of Figure~\ref{fig:architecture}). Concretely, this means: provisioning the runtime services (Test Execution Engine, Collaborative Assessment Dashboard, Report Generator) with the selected plugins; initialising the Testing Database schema; binding role-specific dashboard views and report templates to the configured stakeholder structure; and exposing the resulting environment to the parties of the engagement through RBAC-controlled access. The remainder of this section walks through these runtime components in data-flow order.

\subsection{Shared Data Model and Testing Database}
\label{sub:domain_model}
\textit{Addresses \textbf{R10} (shared data model and persistent storage).}

The shared data model defines the schema through which all assessment results, from both automated tests and manual controls, are harmonised into a single coherent structure. It adopts the \emph{Structured Metrics Metamodel} (SMM) defined by the OMG \cite{OMG_SMM_2018}, extended to accommodate AI-specific concepts: AI system metadata, datasets, evaluation configurations, and legal requirements. These extensions ensure results are stored together with the contextual information necessary for traceability and reproducibility. The model allows a result from a fairness testing tool and a result from a governance controls checklist to be compared, combined, and co-reported within the same representation, and is the technical precondition for any future aggregation of sandbox findings at European level.

The Testing Database is the concrete implementation of the shared data model: it persists harmonised results and exposes them to the rest of the architecture. The reference implementation uses PostgreSQL, but the data model is defined independently of the storage engine, and alternative relational backends could implement it. Portability across such backends is a design intention but has not yet been tested.

\subsection{Test Execution Engine}
\label{sub:execution_engine}
\textit{Addresses \textbf{R7} (role-based access control over test execution); contributes to \textbf{R9} (immutable audit trail) and \textbf{R11} (cybersecurity).}

The Test Execution Engine enables automated, reproducible execution of technical tests with: connectivity to the target AI system's endpoints; versioned per-plugin run configurations for full reproducibility; \ac{RBAC} for fine-grained permission management; isolation of sensitive credentials and target-system endpoints; authenticated inter-component channels; real-time progress tracking; and connection to the Testing Database. Audit logging uses ImmuDB \cite{immudb}, an open-source append-only database that maintains a Merkle hash tree over its contents, allowing any party with the published root hash to detect post-hoc alteration of logged entries; we adopted it over alternative tamper-evident logging approaches because its transactional API is close to that of conventional databases, easing integration. ImmuDB logging is intended to cover every component of the framework, though in the current release it is operational for the Test Execution Engine and the Collaborative Assessment Dashboard only (Section~\ref{sub:limitations}).\footnote{The reference implementation uses Python~3.12 with Django and Django Ninja for type-annotated RESTful APIs, PostgreSQL for persistence, Docker for containerisation, and Celery with RabbitMQ and Redis for asynchronous execution; the web interface uses React and TypeScript, Material UI, TanStack Query, and react-jsonschema-form.}

\subsection{Collaborative Assessment Dashboard}
\label{sub:dashboard}
\textit{Addresses \textbf{R3} (role-specific dashboards for multi-disciplinary interpretation) and \textbf{R7} (role-based access control); contributes to \textbf{R9} (immutable audit trail).}

The Collaborative Assessment Dashboard is the Configurator's primary interface for multi-disciplinary coordination at runtime. It aggregates automated test outcomes and manual controls reviews into a single consolidated, harmonised view, allowing stakeholders to interpret the same underlying evidence through role-appropriate lenses. It provides \ac{RBAC}, interactive visualisations with filtering by dataset, configuration, metric, or test run, structured commenting, integrated report generation, and tamper-evident logging via the same ImmuDB integration described above. The Configurator supports iterative, compliance-by-design workflows: providers can run assessments repeatedly across development cycles, with each iteration guided by expert feedback captured in the commenting system. The webapp is generated using \ac{BESSER} \cite{besser}, a model-driven low-code framework that turns declarative model specifications into running web applications; we chose it because it allows dashboard layouts and role bindings to be regenerated as configuration evolves, rather than hand-coded per engagement. BESSER also provides a no-code drag-and-drop Web Model Editor through which dashboards can be composed visually.

\subsection{Report Generator}
\label{sub:report_generator}

\textit{Addresses \textbf{R8} (automated, audience-segmented report generation).}

The Report Generator produces reports tailored to the plugins used, fully automated once configured. It reads from the Testing Database, injects results into plugin-specific Jinja~2 HTML templates, and renders PDF via Weasyprint. Sections are audience-segmented, with mandatory sections ensuring a minimum level of transparency for all stakeholder groups and optional sections allowing technical granularity to be tuned per audience.

\subsection{Outputs of the Sandbox Process}
\label{sub:outputs}

The sandbox process yields three outputs.
\textbf{AI System Card.} Produced at configuration time but a regulatory artefact in its own right, the AI System Card is a machine-readable, standardised profile of the system, and the basis for the cross-engagement comparability discussed in Section~\ref{sub:qualification}.
\textbf{Tamper-evident Audit Logs} from the Test Execution Engine and the Collaborative Assessment Dashboard, exportable in CSV and JSON.
\textbf{Tailored Assessment Report.} Consolidates the AI System Card, a summary of all tests and controls performed, and for each plugin the dashboard plots with associated stakeholder comments alongside a technical deep-dive. The Tailored Assessment Report is an evidence document produced by the Configurator; it is intended to feed into, or be appended to, the official Exit Report issued by the \ac{CA} at the conclusion of the AIRS engagement, not to replace it.

\section{Pilot}
\label{sec:pilot}

We deployed part of the Configurator within an AIRS pilot run by \ac{EUSAIR}, where the \ac{LIST} contributed as external technical tester. The system under assessment was an \ac{LLM}-backed conversational assistant, designed by an Italian startup to produce \ac{GDPR} \acp{DPIA} through a structured natural-language dialogue in English, French, and Italian. The \ac{EUSAIR} consortium fulfilled the CA-equivalent role for the engagement; stakeholders included a product manager and a developer from the AI provider, a technical tester from LIST, and legal experts from \ac{EUSAIR}.
The pilot exercised the harmonisation and reporting layers of the architecture (the shared data model, the Collaborative Assessment Dashboard, and the Report Generator) under real regulatory conditions. These are the components that determine whether heterogeneous assessment results can be consolidated into coherent, regulator-ready evidence, the core claim of the Configurator. The Test Execution Engine, the Qualification flow, and the end-to-end immutable audit trail were not exercised and remain to be validated in subsequent deployments (Section~\ref{sub:limitations}). We therefore report the pilot as evidence of feasibility for the layers covered, not as confirmation of the architecture as a whole.

Two plugins were used: StrongReject \cite{strongreject_repo, souly2024strongreject_neurips}, an open-source prompt-injection robustness benchmark already in the Catalogue, and a domain-specific plugin targeting cross-model and cross-language \ac{DPIA} completion performance, implemented ad hoc with outputs structured to conform to the shared data model.
The shared data model, the Collaborative Assessment Dashboard, and the Report Generator were used unmodified: no schema extension, dashboard code, or template logic was written for this engagement beyond the plugin-specific report template the integration contract requires of every plugin. StrongReject was reused as released; only the DPIA-completion plugin and its mapping onto the shared data model were built for this engagement.
Both were run outside the Test Execution Engine (exercising the import path rather than end-to-end execution, see Section~\ref{sub:limitations}) and their results ingested via the shared data model \ac{API} as if produced by integrated plugins. That the dashboard and reporting layers operated unchanged on results produced externally is itself evidence of the architecture's modularity: the shared data model, not the execution path, is what binds the components together.

The pilot produced a substantive finding. The initial StrongReject run revealed that the assistant was vulnerable to a non-trivial proportion of the prompt-injection attacks in the benchmark. The product manager and developer added safety guardrails between iterations; a subsequent run showed an observable improvement in jailbreak resistance, a compliance-by-design outcome achieved during the engagement rather than after market deployment. The iterations and the legal experts' interpretation of the residual risk were captured in the Testing Database and propagated through to the Tailored Assessment Report, which was used as an appendix to the official Exit Report produced by \ac{EUSAIR}.

\section{Discussion}
\label{sec:discussion}
This section situates the Configurator beyond the AIRS setting it was designed for, discussing its complementarity with regulatory oversight, the governance of an open Catalogue, the limitations of the current release, and its place in the wider European assessment ecosystem.

\subsection{Complementarity with Regulatory Oversight}
\label{sub:discussion_scope}
While conceived for technical testing within AIRS engagements, the Configurator's open-source modular architecture allows wider use. Providers can employ it for \emph{self-assessment}, running tests and controls iteratively across development cycles to support internal governance and compliance preparation independently of any sandbox engagement. Self-assessment plays a different role from supervised testing: \acp{CA} bring perspectives on operational guidance, dispute resolution, and systemic risks that internal processes are not positioned to provide. Aligning internal practices with regulator-recognised formats can complement supervised engagement by facilitating dialogue, reducing duplication, and strengthening trust in company-led governance.
The same infrastructure provides the technical foundations for \emph{conformity assessment}: reproducible testing, traceability, and harmonised reporting are precisely the properties formal conformity procedures will require. Full operational alignment awaits the finalisation of the relevant implementing acts and harmonised standards \cite{hallensleben2025generative}, but the architecture is designed to accommodate them.

Sandbox effectiveness ultimately depends on what is tested and how results are interpreted. The Configurator can formalise processes, standardise reporting, and provide reusable modules, but it cannot determine the appropriate evaluative focus or weigh findings. The \ac{EUSAIR} pilot illustrated this directly: the same results were interpreted differently by technical, legal, and domain experts. The Configurator complements regulators and legal and compliance specialists by providing structure and transparency; framing and interpretation remain human responsibilities.

\subsection{Catalogue Governance}
\label{sub:discussion_governance}

The Catalogue's three-tier plugin model (Table~\ref{tab:plugin_tiers}) currently operates exclusively at the Core tier: the 10 plugins sit in the main project repository, with the integration contract verified by the project team itself. This concentration is appropriate for the current stage but not sustainable long-term: a Catalogue scaling to the breadth of the \ac{AI Act}'s sectoral and trustworthiness dimensions must admit contributions beyond what a single team can review.
The ambition is to build sufficient traction for a European community to form around the Catalogue, giving open-source assessment-tool projects a concrete incentive to develop plugins and gain visibility through it. Realising that ambition requires opening contribution beyond the project team while preserving the quality signal that makes the Catalogue useful in the first place: a balance between coverage and precision that we are actively working through.
Several properties would need to be vetted before a plugin can be considered for the Verified tier, including basic safety from a cybersecurity perspective and adequate documentation of what the plugin actually measures. Beyond such technical checks, there is a more substantive question: whether vetting should also extend to legal-substantive recognition, so that a plugin's declared mapping to specific legal requirements is endorsed not only by the project team but by an authority competent to make that judgement. The value of such recognition is clear (it would give downstream users, including \acp{CA}, a credible basis for relying on a plugin's claims about which Articles or Annexes it addresses) but the institutional pathway to deliver it is not. Which authorities should play this role, under what mandate, with what resources, and in coordination with which European-level infrastructure initiatives are open questions that will require broader discussion among \acp{CA}, EU-funded initiatives, and the assessment-tool community itself.
The Community tier raises a different question: how to surface a long tail of contributions for discoverability without implying endorsement, and how to signal that distinction clearly enough that downstream users, including those reading reports that cite a plugin, can tell at a glance whether a result rests on a vetted tool or an unvetted one. We have not settled on the right design here either.

\subsection{Limitations and Roadmap}
\label{sub:limitations}

The paper presents an architecture validated through an early-stage pilot covering the harmonisation and reporting layers; broader claims (Section~\ref{sub:discussion_ecosystem}) are prospective and contingent on the roadmap items below.
The AI System Card generator has not been evaluated systematically: manual amendment is available and has been used to correct the failure modes described in Section~\ref{sub:qualification}, but subtle inconsistencies may go undetected if the amendment step is skipped, and quantifying run-to-run consistency is a near-term priority alongside the agentic replacement.
The Catalogue is currently available as a web-only public instance. Organisations with confidentiality constraints cannot integrate proprietary plugins under this model; a downloadable open-source local-instance release is planned to address this. Test selection remains a manual process of Catalogue navigation, imposing a non-trivial burden on users unfamiliar with the assessment landscape; a strictly advisory recommendation wizard, combining ontology-based matching with agentic reasoning and leaving the selection decision with the user, is on the roadmap.

The shared data model has been validated against 10 plugins across two trustworthiness domains, and its generalisability to specialised domains or novel system types has not yet been demonstrated. Validation will continue as new plugins are admitted through the Verified and Community tiers. Database portability beyond the PostgreSQL reference implementation remains a design intention that has not yet been tested against alternative relational backends.
The immutable audit trail is operational for the Test Execution Engine and the Collaborative Assessment Dashboard only; no audit logs were generated in the pilot. Extending ImmuDB logging to all components is a near-term priority and will be exercised in subsequent pilots. Conformity assessment support remains at the level of technical foundations; full alignment with formal conformity workflows awaits the finalisation of the relevant implementing acts and harmonised standards \cite{hallensleben2025generative}.

\subsection{Harmonisation with other EU initiatives}
\label{sub:discussion_ecosystem}

The shared data model and harmonised reporting have implications beyond individual sandbox engagements, though we treat these as prospective rather than demonstrated. If widely adopted across Member States, the structured, machine-readable outputs the Configurator produces could enable comparative analysis of assessment findings at European scale, a precondition for the evidence-based regulatory learning Article~58(2)(i) anticipates and for the forward-looking governance functions \acp{AIRS} are designed to support \cite{ahernNewAnticipatoryGovernance2025}. The fragmentation of assessment tools and output formats is a structural barrier to the accumulation of governance knowledge \cite{schiff2025strategies}. A shared infrastructure lowers that barrier without imposing uniformity, since the plug-in architecture preserves the diversity of testing approaches while the data model ensures their outputs remain interoperable.
National guidelines, sector-specific requirements, and bespoke evaluation tools can be accommodated alongside shared test types, templates, and expert modules. As noted in prior work \cite{due2024sandboxing}, knowledge-sharing mechanisms are critical to avoid duplication and support harmonisation while preserving local flexibility. European initiatives such as \acp{EDIH} \cite{eu_edihs}, AI Factories \cite{ai_factories_2025}, and \acp{TEF} \cite{ec:tefs}, whose importance for \acp{AIRS} is acknowledged in the current draft of the Implementing Act \cite{implementingAct}, illustrate institutional environments where such an approach could be explored.

A further integration point is the \ac{AIoD} Platform \cite{aiod_platform_2023}. The Configurator is a natural candidate for registration on \ac{AIoD} as an assessment service, but the more consequential integration runs in the opposite direction: using the Configurator to assess models and datasets \emph{published} on \ac{AIoD}, closing the gap between AIoD's current state (where resources are listed but not systematically assessed for trustworthiness) and what end-users would need to rely on those resources with confidence. The vision of a European sovereign open-source asset for AI assessment is a long-term aspiration; the contribution of this paper is the architectural and governance groundwork on which such an asset could be built.

\section{Conclusion}
\label{sec:conclusion}

This paper presented the AI Assessment Sandbox Configurator, an open-source framework that supports technical testing within \acp{AIRS} under the EU \ac{AI Act} by assembling a bespoke assessment environment for each engagement. From the procedural conditions of Articles 57 and 58 and consultations with Competent Authorities, we derived 11 architectural and governance requirements, and met them with an architecture built around two stable interfaces --- the Catalogue plug-in API and the shared data model --- with all other components substitutable. We reported an early-stage pilot conducted within an \ac{EUSAIR} AIRS engagement that exercised the harmonisation and reporting layers and produced an observable improvement in the assessed system.
Beyond its immediate European context, the framework is intentionally transferable: the separation between regulatory objectives (\acp{AIRS}), the need for technical testing evidence, and the infrastructure used to produce it reflects a general governance pattern relevant to other jurisdictions exploring AI oversight \cite{moraes2025regulatory, nabil2024artificial}. Future work, detailed in Section~\ref{sub:limitations}, will deliver the roadmap components, validate the shared data model against a broader range of plugins, exercise the full end-to-end workflow in subsequent pilots, and grow the contributor community. If sandboxes are to serve as instruments of regulatory learning rather than isolated compliance exercises, the assessments conducted within them must produce evidence that is comparable, traceable, and reusable across organisations, sectors, and Member States.

\section*{Acknowledgments}
We thank the Luxembourg AI Factory (L-AIF, grant agreement ID 1012343366) for supporting this work.


\bibliography{aaai2026}
\end{document}